\pdfoutput=1

\documentclass[11pt]{article}

\usepackage{ACL2023}

\usepackage{times}
\usepackage{latexsym}
\usepackage{booktabs}

\usepackage[T1]{fontenc}

\usepackage[utf8]{inputenc}

\usepackage{microtype}

\usepackage{inconsolata}

\title{HausaNLP at SemEval-2025 Task 3: Towards a Fine-Grained Model-Aware Hallucination Detection}

\author{
Maryam Bala$^{1}$, Amina Imam Abubakar$^{1,2}$, 
\textbf{Abdulhamid Abubakar}$^1$, \\ \textbf{Abdulkadir Shehu Bichi}$^1$, \textbf{Hafsa Kabir Ahmad}$^{1,3}$, \textbf{Sani Abdullahi Sani}$^{1}$, \\ \textbf{Idris Abdulmumin}$^{1,4}$, \textbf{Shamsuddeen Hassan Muhamad}$^{1,3,5}$, \textbf{Ibrahim Said Ahmad}$^{1,3,6}$ \\\footnotesize{$^{1}$HausaNLP, $^{2}$University of Abuja, $^{3}$Bayero University Kano, $^{4}$Data Science for Social Impact, University of Pretoria,} \\ \footnotesize{$^{5}$Imperial College London, $^6$Northeastern University} \\
\footnotesize{\texttt{\textbf{correspondence}: maryam.bala@outlook.com, i.ahmad@northeastern.edu}
}
}

\begin{document}
\maketitle
\begin{abstract}

This paper presents our findings of the Multilingual Shared Task on Hallucinations and Related Observable Overgeneration Mistakes, MU-SHROOM, which focuses on identifying hallucinations and related overgeneration errors in large language models (LLMs). The shared task involves detecting specific text spans that constitute hallucinations in the outputs generated by LLMs in 14 languages. To address this task, we aim to provide a nuanced, model-aware understanding of hallucination occurrences and severity in English. We used natural language inference and fine-tuned a ModernBERT model using a synthetic dataset of 400 samples, achieving an Intersection over Union (IoU) score of 0.032 and a correlation score of 0.422. These results indicate a moderately positive correlation between the model's confidence scores and the actual presence of hallucinations. The IoU score indicates that our model
has a relatively low overlap between the predicted hallucination span and the truth annotation. The performance is unsurprising, given the intricate nature of hallucination detection. Hallucinations often manifest subtly, relying on context, making pinpointing their exact boundaries formidable.
\end{abstract}

\section{Introduction}
Despite the advancements in Natural Language Processing (NLP) and the development of Natural Language Generation (NLG) models, their limitations and potential risks have gained increased attention. A significant issue is that NLG models often produce unfaithful text relative to the source input, a phenomenon known as "hallucination" \cite{3koehn2017six, 4ohrbach2018object, 5aynez2020faithfulness}. Hallucinations in NLG models often result in outputs that, while fluent, lack accuracy. This issue arises because existing evaluation metrics prioritize fluency over correctness, ultimately diminishing system performance and failing to meet user expectations in practical applications \cite{6mickus2024semeval}. For example, \citet{16dopierre2021protaugment} illustrate this phenomenon by attempting to paraphrase the statement "I am not sure where my phone is," which leads to the hallucinated output: "How can I find the location of any Android mobile." 

Hallucination in NLG is concerning due to its impact on performance and safety in applications like medicine, where hallucinatory summaries or machine-translated instructions can pose risks to patient diagnosis \cite{12023survey}. Similarly, hallucinations can lead to privacy violations by generating sensitive information not present in the source input \cite{2carlini2021extracting}.

To address this challenge, \textit{“SemEval-2025 Task 3 – Mu-SHROOM, a Shared-task on Hallucinations and Related Observable Overgeneration Mistakes”} \cite{vazquez-etal-2025-mu-shroom} was introduced. The shared task involves detecting specific text spans that constitute hallucinations in outputs generated by LLMs. Participants compute the probability of each character being marked as a hallucination using the LLM output consisting of a character string, tokens, and logits, thus enabling a fine-grained hallucination detection.

\section{Background}
Numerous efforts are provided to address hallucination across various NLG tasks. Analyzing hallucinatory content and its relationships in different tasks could enhance our understanding and unify efforts across NLG fields \cite{12023survey}. While most existing studies focus on specific tasks like abstractive summarization \cite{7uang2021factual,8maynez2020faithfulness} and machine translation \cite{9lee2018hallucinations}, the study of \citet{12023survey} offer a comprehensive analysis on the phenomenon of hallucination in  abstractive summarization, dialogue generation, generative question answering, data- to-text generation, and machine translation.
\newline
\newline
\cite{10parikh2020totto} argue that hallucination problem occurs  when there is very little divergence in dataset and encoder with a defective comprehension ability could influence the degree of hallucination. Similarly, \cite{11koehn2017six} show that training and modeling choices of neural models have influence for hallucination. While large pre-trained models used for downstream NLG tasks are powerful in providing generalizability and coverage, they however prioritize parametric knowledge over the provided input and can result in hallucination of excess information in the output \cite{12longpre2021entity}. 

Recent efforts to address hallucination include the Shared Task on Hallucinations and Related Observable Overgeneration Mistakes (SHROOM), which focused on binary classification for task-specific English language models \cite{6mickus2024semeval}. \citet{13maksimov2024deeppavlov,14arzt2024tu,15rahimi2024hallusafe} identify cases of fluent overgeneration hallucinations in model-aware and model-agnostic settings. They detect grammatically sound outputs which contain incorrect or unsupported semantic information. Building on this task instead of focusing on model-agnostic and model-aware tracks, this year’s task focuses on the multilingual aspect. Therefore, all data-points in this year’s task are model-aware.
 
We present our Multilingual Shared-task on Hallucinations and Related Observable Overgeneration Mistakes (Mu-SHROOM) which aims to detect hallucinations in a multilingual context, providing a more nuanced understanding of their occurrence and severity.
MU-SHROOM is a SemEval-2025 shared task that focuses on detecting hallucinations and overgeneration errors in AI-generated text, a crucial challenge in improving the reliability of LLMs \cite{vazquez-etal-2025-mu-shroom}. Hallucinations occur when LLMs produces fluent but false or unsupported content, while overgeneration mistakes involve excessive, often misleading text \cite{12023survey}. This task is important as language models become increasingly prevalent in various applications, where factual accuracy and reliability are essential for maintaining user trust and system integrity. Therefore, tackling these issues is essential for ensuring AI-generated text remains trustworthy and useful across various applications \cite{13maksimov2024deeppavlov}.

The task consists of participants detecting spans of text corresponding to hallucinations and determine which parts of the given text produced by LLMs constitute hallucinations. Annotated dataset are provided, allowing researchers to develop and benchmark models for identifying these issues in different linguistic contexts. Given the LLM output as a string of characters, a list of tokens, and a list of logits, participants  calculate the probability that each character is marked as a hallucination and thus provide a fine- grained hallucination detection. Similarly, the task is held in multilingual and multi-model context as data are produced by a variety of public-weights LLM in multiple languages which includes: Arabic (Modern standard), Chinese (Mandarin), English, Finnish, French, German, Hindi, Italian, Spanish, and Swedish, along with three surprise test languages.

\section{System Overview}
One of the primary challenges in this task was the lack of labeled training data. To address this, we created a synthetic training dataset using ChatGPT, adhering to official annotation guidelines. 

For the language selection, we used English language only and this is due to time constraint. Secondly, English language offers an extensive resource for NLP model development and evaluation. Similarly, focusing on a single language allowed us to develop deeper linguistic pattern recognition for hallucination detection.  

For model selection, we fine-tuned ModernBERT \cite{modernbert}, an advanced variant of BERT \cite{devlin-etal-2019-bert} that offers significant improvements in handling long-context inputs, computational efficiency, and robustness in token classification tasks. This choice aligns with recent studies highlighting the benefits of long-context language models in detecting nuanced text errors \cite{sahitaj2025automatedfactcheckingrealworldclaims}.

For a precise understanding of hallucination detection, we used the model-aware method. This method is based on analysing internal data of LLM during inference. One of the possible approaches is the analysis of the outputs of the hidden layers of the transformer. Using vector values of hidden layers for hallucination detection was proposed in a method called Statement Accuracy Prediction, based on Language Model Activations (SAPLMA) \cite{17azaria2023internal}. SAPLMA is a probing technique that utilises a feedforward neural network trained on activation values of the hidden layers of LLM.

	
	
	
	

\section{Experimental Setup}

\subsection{Dataset Generation}

For this task, we initiated the development of a robust model for detecting hallucinations in text by generating synthetic data specifically for training purposes. \citet{borra-etal-2024-malto} have shown the success of using synthetic data in finetuning models for hallucination detection in LLMs. This synthetic dataset was designed to simulate a wide range of scenarios, enhancing the model's ability to generalize across various contexts. With this approach, We were able to generate 400 diverse labeled data points. 

Similarly, we utilized the task-provided validation dataset to assess the model's performance during the training phase, ensuring that it effectively learned from the training data while maintaining its ability to generalize. Finally, we reserved the provided unlabeled test dataset for final evaluation, allowing us to measure the model's performance on completely unseen data.

\subsection{Data Preprocessing}
The text data was tokenized using the Hugging Face AutoTokenizer from the Transformers library \citep{wolf-etal-2020-transformers}, which efficiently segmented the text into manageable units. Similarly, token labels were assigned based on entity spans identified in the dataset, ensuring that the model learned to recognize specific entities and their contexts. 

\subsection{Model Training}
The core of our approach involved fine-tuning the ModernBert model on the synthetic dataset. During this training process, we implemented several advanced techniques to optimize performance:
\begin{itemize}
    \item Cosine Learning Rate Scheduler: This scheduler dynamically adjust the learning rate throughout the training, promoting a smoother convergence and help avoid local minima.
    \item Mixed Precision Training: By utilizing both float16 and float32 data types, we enhanced computational efficiency while preserving model accuracy. This approach significantly reduced memory usage and improved training speed.
    \item Gradient Clipping: To maintain stability during training, we employed gradient clipping techniques that prevented gradients from exceeding a specified threshold. This safeguard helped mitigate issues related to exploding gradients, which can destabilize the training process.
\end{itemize}

\subsection{Evaluation Metrics}
To evaluate the model's effectiveness post-training, we utilized several key metrics: precision, recall, and F1 score. The selected evaluation metrics were implemented to facilitate a comprehensive assessment of the model’s efficacy, encompassing both predictive precision and capacity to address class distribution asymmetries. The above metrics are used in addition to Intersection over Union (IoU) and Correlation Score that were used by the task organizers to assess the performance of our model.

\subsection{Classification and Prediction}
Once trained, the model was capable of processing unseen text to detect hallucinations effectively. The outputs of the model were converted into two formats: hard labels (binary classification) indicating the presence or absence of hallucinations, and soft labels representing confidence scores that quantify the model's certainty regarding its predictions. This dual-output approach not only enhances interpretability but also allows for flexible integration into downstream applications where varying levels of confidence may be required.

	
	
	
	

\section{Results}
At the end of training our model, the evaluation result showed that the model had a precision and recall score of 0.49 and 0.54 respectively. The F1 score of the model was at 0.43. The result indicates that the model correctly identifies hallucinations about half the time. The model is also able to detect slightly more than half of all hallucinations present in the text.
\par On the task-based evaluation of our model, our system achieved an Intersection over Union (IoU) score of 0.032 and a correlation score of 0.422. The IoU score indicates that our model has a relatively low overlap between the predicted hallucination span and the truth annotation. With the relatively low score, the model is struggling to identify the exact boundaries of the hallucinated content. Going by this result also, the model may be prone to false positives and/or false negatives.
\par The models correlation score of 0.422 shows a moderate positive correlation between the confidence scores of the model and the actual presence of hallucinations. This result can translate to the model's ability to differentiate between hallucinated and non-hallucinated content. While this may not be the best performance, the results show that there is room for further improvement.
\par Our model results ranked 42nd and 24th on the IoU and correlation score indices respectively. The scores are not favourable and may not be entirely unexpected given the inherent complexity of hallucination detection,. Hallucinations can be subtle and context-dependent, making exact boundary detection particularly challenging. The model result is a promising starting point for further improvement.
	
	
	

\section{Conclusion}
This paper presents our approach to the SemEval shared task on LLM hallucination detection, focusing on a fine-grained, model-aware analysis of hallucination occurrences and severity in the English language. We leveraged natural language inference and fine-tuned a ModernBERT model using a synthetic dataset of 400 samples. Our model achieved rankings of 42nd and 24th on the Intersection over Union (IoU) and correlation score indices, respectively. These results, while modest, underscore the inherent complexity of hallucination detection and highlights the need for continued refinement and innovation in this area. Our findings serve as a promising foundation for future improvements, emphasizing the importance of model-aware strategies in enhancing the reliability of LLM outputs.

\bibliography{custom}
\bibliographystyle{acl_natbib}



\end{document}